\documentclass[a4paper,12pt]{article}
\usepackage[utf8]{inputenc}
\usepackage{threeparttable}
\pdfoutput=1
\usepackage{verbatim}
\usepackage{epsfig}
\usepackage{epsf}
\usepackage{epstopdf}
\usepackage{graphicx}
\usepackage{textcomp}
 \usepackage{amsmath}
\usepackage{amssymb}
\usepackage{color}
\definecolor{Blue}{rgb}{0.3,0.3,0.9}
\definecolor{red}{rgb}{1,0,0}
\usepackage{rotating}
\usepackage{pgfplots}
\usepackage{tikz}
\usepackage{url}
\usetikzlibrary{patterns}
\usepackage{amsfonts,amsthm,titling,url,array,mathtools}
\usepackage{cite}
\usepackage[toc,acronym]{glossaries}

\title{Is `Unsupervised Learning' a Misconceived Term?}
\author{Stephen G. Odaibo,\\ M.D.,M.S.(Math),M.S.(Comp. Sci.)\\{\tiny{.}}\\\vspace{10pt}RETINA-AI Health, Inc.}

\begin{document}

\maketitle
\thispagestyle{empty}
\begin{abstract}
Is all of machine learning supervised to some degree? The field of machine learning has traditionally been categorized pedagogically into \textit{supervised} vs \textit{unsupervised learning}; where supervised learning has typically referred to learning from labeled data, while unsupervised learning has typically referred to learning from unlabeled data. In this paper, we assert that all machine learning is in fact supervised to some degree, and that the scope of supervision is necessarily commensurate to the scope of learning potential. In particular, we argue that clustering algorithms such as k-means, and dimensionality reduction algorithms such as principal component analysis, variational autoencoders, and deep belief networks are each internally supervised by the data itself to learn their respective representations of its features. Furthermore, these algorithms are not capable of external inference until their respective outputs (clusters, principal components, or representation codes) have been identified and externally labeled in effect. As such, they do not suffice as examples of unsupervised learning. We propose that the categorization `supervised vs unsupervised learning' be dispensed with, and instead, learning algorithms be categorized as either \textit{internally or externally supervised} (or both). We believe this change in perspective will yield new fundamental insights into the structure and character of data and of learning algorithms.

\vspace{25pt}{\hspace{-17pt}\small{\textbf{Correspondence Email:}\\\hspace{10pt} stephen.odaibo@retina-ai.com}}
\end{abstract}

\newpage
\section{Introduction}

\begin{figure}[h]
\begin{center}
\scalebox{.4}
{\includegraphics{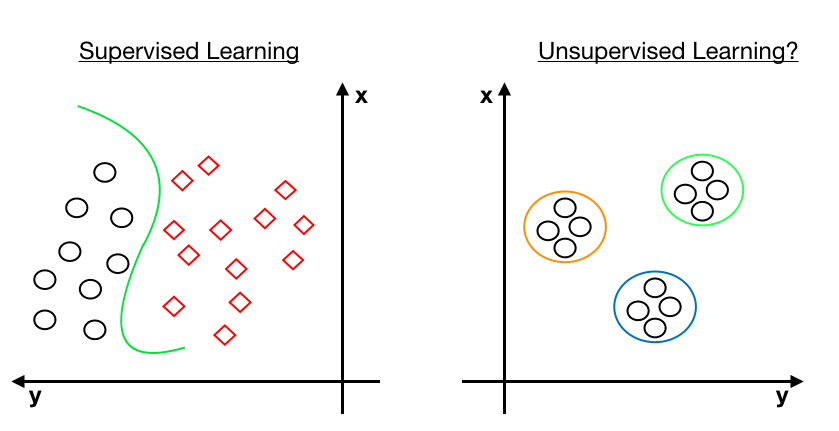}}
\end{center}
\caption[SupvsUnsup]{Prototypical Illustration of Supervised vs Unsupervised Learning}
\label{fig:SupvsUnsup}
\end{figure}

Traditional thinking has been to pedagogically divide machine learning into supervised vs unsupervised learning, as depicted in Figure(\ref{fig:SupvsUnsup}). In this paper, we challenge that scheme by arguing that all machine learning is supervised to some degree. In the literature, the methods which have typically been referred to as \textit{unsupervised learning} include methods such as kmeans\cite{ma1967,waca2001,cong2011,cong2012} which cluster data based on some distance metric, and methods which attempt to derive some representation code of the data in terms of the algorithm's individual architecture. This latter class of methods can be thought of on the one hand as dimensionality reducers, and they include autoencoders\cite{lera2011,hisa2006,legr2009,vila2010,ngkh2011}, principal component analysis\cite{jo2011}, deep belief networks\cite{hios2006}, and adversarial generative models such as predictability minimization\cite{sc1992} and generative adversarial networks\cite{gopo2014,rame2015,mios2014}. Another interpretation for these is as density estimators which effectively maximize the likelihood of the input data. In other words, they attempt to derive a distribution estimate, $Q(x)$, for the native probability distribution, $P(x)$, of the data, $\chi$, by minimizing the Kullback-Liebler or related distance:

\begin{equation}
 D_{KL}(P||Q) = -\sum_{x\in \chi} P(x)\log \left(\frac{Q(x)}{P(x)}\right)
\end{equation}

Each of the above mentioned algorithms are guided strongly by the data, which is in every sense their supervisor. In the case of kmeans clustering, the location of the data point is the label and the loss is its distance to the k centroids. In the case of deep neural autoencoders, the input image is the label and the loss is a function of the distance between the original and the reconstructed image. And in the case of principal component analysis, the multivariate data points are the labels, and they contain all the information needed to select the mutually orthogonal set of principal component vectors.

In the above examples, this direct supervision by the data itself constrains the learning potential to things intrinsic to the data, e.g. internal feature representations of the data. If on the other hand, one wanted to learn things externally ascribed to the data, i.e labels, then external supervision via implicit or explicit labeling will be necessary. Prior to such labeling, these algorithms are not directly useful for inference of externally defined meaning. For example, without some form of feature identification and labeling, an image autoencoder's output cannot be used to directly infer things such as `cat,' `dog,' or `face.' Therefore one must recognize these methods as supervised learning algorithms where the supervision is by the data itself and the learning is of a representation of the data's internal features.

We propose that the categorization `supervised vs unsupervised learning' be dispensed with, and instead, learning algorithms be categorized as either \textit{internally or externally supervised} (or both). We believe this change in perspective will yield new fundamental insights into the structure and character of data and of learning algorithms.

\begin{figure}[h]
\begin{center}
\scalebox{.46}
{\includegraphics{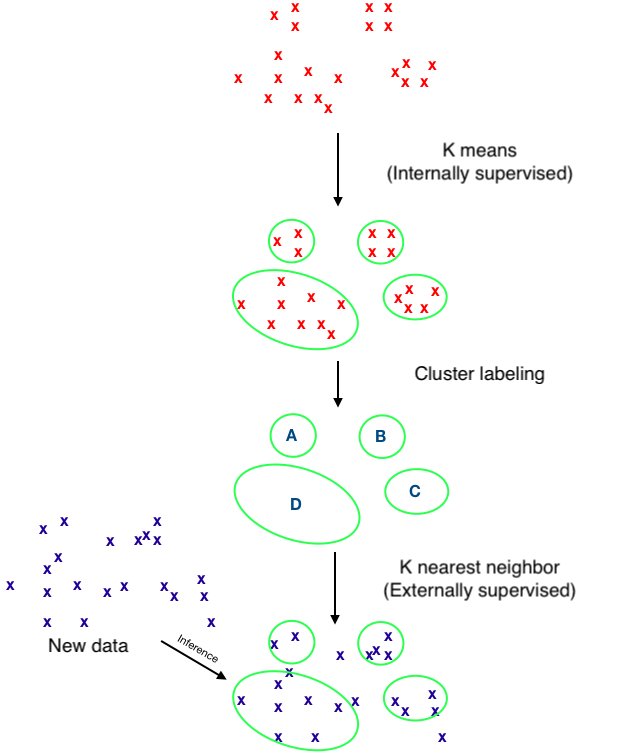}}
\end{center}
\caption[kmeans to kNN]{Kmeans-to-kNN: Internal-to-External Supervision}
\label{fig:KtK}
\end{figure}

\section{Supervision, Learning, and Inference Scope}

Here, we propose the concept of scope in regards to supervision, learning, and inference. We conjecture that all machine learning is supervised to some degree. And the level and scope of supervision determines the level and scope of learning potential. This in turn determines the level and scope of inference potential of which the algorithm's output is capable. An algorithm cannot learn beyond the scope of its supervision, and cannot infer beyond the scope of what it has learned. For instance, when the only supervisor is the data itself and there is no external supervision in the form of labels, such an algorithm will only be capable of learning things intrinsic to the data; things such as internal feature representation codes. By the same token, the inference which such an algorithm will be capable of will be restricted to what it has learned, for example it will be able to represent new unseen data in the learned representation code. This can be considered an intrinsic data-level scope of supervision, learning, and inference. On the other hand, if externally defined meaning is ascribed to the learned features codes or some algebraic combination of them like an image, then the scope of learning and inference will similarly go beyond intrinsic features of the data. In this instance, there will be a mapping into the corresponding semantic space.

\section{Data as Supervisor}
When there are no labels, i.e. no external supervision to guide the learning process, one is limited to internal supervision by the data itself. Under this circumstance, the range of admissible learning processes can be summarized as \textit{basis feature selection}. This may include a change of basis feature operation; or a reduction in span, i.e. a dimensionality reduction. The entire range of data-supervised examples such as kmeans clustering, principal component analysis, autoencoding, deep belief nets, and generative adversarial nets can all be described as basis feature selection operations. The possible inference in this case is essentially limited to expressing new data in the learned feature basis.

\section{Some Examples}

Table \ref{tab:category} shows some examples of algorithms and their associated supervision type. The table assumes the standard implementation of the algorithms. The previously mentioned examples where the data itself is the supervisor are all internally supervised. The examples on the table are externally supervised. Notably, the external labeling is not always obvious. GANs are such an example. For most machine learning applications, some form of external inference capability is ultimately desired. To enable this, the internally supervised output will need to be converted to externally supervised method as exemplified by kmeans-to-kNN illustrated in Figure~(\ref{fig:KtK}). In such cases, the composite process is therefore both internally and externally supervised. 

\begin{table}
\begin{small}
\begin{center}
\begin{tabular}{|c|c|}
\hline
\textbf{Algorithm} & \textbf{Supervision} \\
\hline\hline
Kmeans clustering & Internal\\
Variational autoencoders & Internal\\
Deep belief networks & Internal\\
Principal component analysis & Internal\\
K nearest neighbor (kNN) & External\\
CNN image classification & External\\
Generative adversarial nets & External\\
Kmeans-to-kNN & Both\\
\hline
\end{tabular}
\end{center}
\end{small}
\caption[category]{Supervision type of some algorithms}
\label{tab:category}
\end{table}

\subsection{GANs are Externally Supervised}
Generative adversarial networks (GANs) are an interesting example of an algorithm that is surprisingly \textit{externally supervised}, but have often been called `unsupervised'\cite{rame2015}. Their individual modules such as the discriminator or the generator considered in isolation, are externally supervised. The Discriminator in a GAN is a binary classification convolutional neural network with clearly labeled outputs: {real vs synthesized}; real if the image source is the training data and synthesized if the image source is the Generator. Similarly, the Generator is a deconvolutional neural network whose weights are updated based on the Discriminator's assessment of the Generator's clearly labeled output. This clearly labeled output are what directly constitute the loss function and guide the training, as exemplified by the stochastic gradient of the Shannon-Jensen loss:

\begin{equation}
 \nabla_{{\theta}_d}\frac{1}{m}\displaystyle\sum_{i=1}^m\big[\log D\left(x^{(i)}\right)+\log\left(1-D\left(G\left(z^{(i)}\right)\right)\right)\big]
\end{equation}

 The entire process of training a GAN is guided by the algorithm designer's knowledge and labeling of the training data as \textit{real} and the generator's output as \textit{synthesized}.

\section{Discussion}

No algorithms in existence are truly unsupervised. Instead, the scope of supervision of algorithms determine the learning potential, which in turn determines the inference potential of resulting trained model. We found that most algorithms that have been termed `unsupervised' in the literature are supervised by the data, i.e. they they are internally supervised. The learning potential of such data-level scope algorithms is to determine a `basis' feature set to represent the data. The specific mechanisms of data-level scope supervision varied between algorithms, but each presented the data as the standard against which the resulting basis feature set was evaluated.

An algorithm can be termed unsupervised if it is able to learn outside of the scope of its supervision. We however conjecture that this is impossible. For instance, while there is a representation for face, dog, cat, or any other visual object in terms of any appropriately deep neural network architecture, one does not have semblance of intelligence till the representations are mapped to meaning. In other words, one can not make useful and meaningful inference until the derived basis features are mapped to meaningful objects such as `face', `cat', and `dog'. This is labeling and makes it an externally supervised problem.

Neural networks learn so many more features that we can currently track or be aware of. For instance, in a standard convolutional neural network image classification problem, our labels are based on certain salient discriminative features, yet the network may utilize other and more (or less) features for discrimination. This makes it clear that the designation of supervision type is based on the intended task. For instance, assigning class labels in an image classification problem suffices to designate the algorithm as `supervised' regardless of how exactly the neural network makes its determination, which may be via a feature different from the label. By the same token, when the task is to find an internal representation of a dataset, then presenting the algorithm with a representative sample of that dataset clearly qualifies as supervision, and in this work we have termed such supervision \textit{internal}. This is in contrast to \textit{external supervision} in which the internal `basis' feature are mapped to an external label.

In contrast to previous thinking which categorized supervision as either present (supervised) or absent (unsupervised); in this work, we have argued that supervision is always present in learning algorithms and is either \textit{internal}, \textit{external}, or both. We have also pointed out how the scope of supervision determines the scope of learning potential, which in turn determines the scope of inference potential. We anticipate that this change in perspective will yield new fundamental insights into the structure and character of data and of learning algorithms.

\section*{Acknowledgement}
I initially posted this idea on Linkedin on Friday March 29th 2019, where it generated good discussion. I'm thankful to discussion participants such as Emml Asimadi, Adebayo Aderibigbe, Idris Azeez, Busayo Coker, Oden Vangelis, and Busayo Olukunle amongst others. 

\bibliographystyle{apalike}
\bibliography{/home/sodaibo/Documents/BOOKS/mybibliography_2015_nov}

\begin{thebibliography}{}

\bibitem[Coates et~al., 2011]{cong2011}
Coates, A., Ng, A., and Lee, H. (2011).
\newblock An analysis of single-layer networks in unsupervised feature
  learning.
\newblock In {\em Proceedings of the fourteenth international conference on
  artificial intelligence and statistics}, pages 215--223.

\bibitem[Coates and Ng, 2012]{cong2012}
Coates, A. and Ng, A.~Y. (2012).
\newblock Learning feature representations with k-means.
\newblock In {\em Neural networks: Tricks of the trade}, pages 561--580.
  Springer.

\bibitem[Goodfellow et~al., 2014]{gopo2014}
Goodfellow, I., Pouget-Abadie, J., Mirza, M., Xu, B., Warde-Farley, D., Ozair,
  S., Courville, A., and Bengio, Y. (2014).
\newblock Generative adversarial nets.
\newblock In {\em Advances in neural information processing systems}, pages
  2672--2680.

\bibitem[Hinton et~al., 2006]{hios2006}
Hinton, G.~E., Osindero, S., and Teh, Y.-W. (2006).
\newblock A fast learning algorithm for deep belief nets.
\newblock {\em Neural computation}, 18(7):1527--1554.

\bibitem[Hinton and Salakhutdinov, 2006]{hisa2006}
Hinton, G.~E. and Salakhutdinov, R.~R. (2006).
\newblock Reducing the dimensionality of data with neural networks.
\newblock {\em science}, 313(5786):504--507.

\bibitem[Jolliffe, 2011]{jo2011}
Jolliffe, I. (2011).
\newblock {\em Principal component analysis}.
\newblock Springer.

\bibitem[Le et~al., 2011]{lera2011}
Le, Q.~V., Ranzato, M., Monga, R., Devin, M., Chen, K., Corrado, G.~S., Dean,
  J., and Ng, A.~Y. (2011).
\newblock Building high-level features using large scale unsupervised learning.
\newblock {\em arXiv preprint arXiv:1112.6209}.

\bibitem[Lee et~al., 2009]{legr2009}
Lee, H., Grosse, R., Ranganath, R., and Ng, A.~Y. (2009).
\newblock Convolutional deep belief networks for scalable unsupervised learning
  of hierarchical representations.
\newblock In {\em Proceedings of the 26th annual international conference on
  machine learning}, pages 609--616. ACM.

\bibitem[MacQueen et~al., 1967]{ma1967}
MacQueen, J. et~al. (1967).
\newblock Some methods for classification and analysis of multivariate
  observations.
\newblock In {\em Proceedings of the fifth Berkeley symposium on mathematical
  statistics and probability}, volume~1, pages 281--297. Oakland, CA, USA.

\bibitem[Mirza and Osindero, 2014]{mios2014}
Mirza, M. and Osindero, S. (2014).
\newblock Conditional generative adversarial nets.
\newblock {\em arXiv preprint arXiv:1411.1784}.

\bibitem[Ngiam et~al., 2011]{ngkh2011}
Ngiam, J., Khosla, A., Kim, M., Nam, J., Lee, H., and Ng, A.~Y. (2011).
\newblock Multimodal deep learning.
\newblock In {\em Proceedings of the 28th international conference on machine
  learning (ICML-11)}, pages 689--696.

\bibitem[Radford et~al., 2015]{rame2015}
Radford, A., Metz, L., and Chintala, S. (2015).
\newblock Unsupervised representation learning with deep convolutional
  generative adversarial networks.
\newblock {\em arXiv preprint arXiv:1511.06434}.

\bibitem[Schmidhuber, 1992]{sc1992}
Schmidhuber, J. (1992).
\newblock Learning factorial codes by predictability minimization.
\newblock {\em Neural Computation}, 4(6):863--879.

\bibitem[Vincent et~al., 2010]{vila2010}
Vincent, P., Larochelle, H., Lajoie, I., Bengio, Y., and Manzagol, P.-A.
  (2010).
\newblock Stacked denoising autoencoders: Learning useful representations in a
  deep network with a local denoising criterion.
\newblock {\em Journal of machine learning research}, 11(Dec):3371--3408.

\bibitem[Wagstaff et~al., 2001]{waca2001}
Wagstaff, K., Cardie, C., Rogers, S., Schr{\"o}dl, S., et~al. (2001).
\newblock Constrained k-means clustering with background knowledge.
\newblock In {\em Icml}, volume~1, pages 577--584.

\end{thebibliography}

\begin{figure}[h]
\begin{center}
\scalebox{.40}
{\includegraphics*{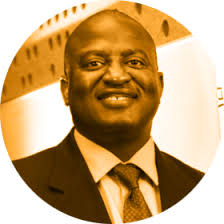}}
\end{center}
\caption{\textbf{About the Author:}  Dr. Stephen G. Odaibo is Founder, CEO, and Chief Software Architect of RETINA-AI Health Inc, a company using Artificial Intelligence to improve Healthcare. He is a Retina specialist, Mathematician, Computer Scientist, and Full-Stack AI Engineer. Dr. Odaibo is the only Ophthalmologist in the world with advanced degrees in both Mathematics and Computer Science. In 2017 UAB College of Arts and Sciences awarded Dr. Odaibo its highest honor, the Distinguished Alumni Achievement Award. In 2005 he won the Barrie Hurwitz Award for Excellence in Clinical Neurology at Duke Univ. School of Medicine where he topped the class in Neurology and in Pediatrics. In 2016 Dr. Odaibo delivered the Opening Keynote address at the Global Ophthalmologists Meeting in Osaka Japan. And he delivered the inaugural Special Guest Lecture in Ophthalmology at the University of Ilorin, Nigeria. In 2018, Dr. Odaibo delivered the keynote address at the National Medical Association's New Innovations in Ophthalmology Session. And he delivered a Plenary Keynote address on AI in Healthcare at AI Expo Africa in Cape town, South Africa. He is author of the book "Quantum Mechanics and the MRI Machine'' (2012), and of the book "The Form of Finite Groups: A Course on Finite Group Theory" (2016). Clinically, Dr. Odaibo focuses on caring for patients with macular degeneration, diabetic retinopathy, retinal vascular occlusions, retinal tears, and localized retinal detachments.}
\label{fig:auth}
\end{figure}

\end{document}